\documentclass{article}

\usepackage{PRIMEarxiv}

\usepackage[utf8]{inputenc} 
\usepackage[T1]{fontenc}    
\usepackage{hyperref}       
\usepackage{url}            
\usepackage{booktabs}       
\usepackage{amsfonts}       
\usepackage{nicefrac}       
\usepackage{microtype}      
\usepackage{lipsum}
\usepackage{fancyhdr}       
\usepackage{graphicx}       
\graphicspath{{media/}}     

\usepackage{amsmath,amssymb,amsfonts}
\usepackage{algorithmic}
\usepackage{graphicx}
\usepackage{textcomp}
\usepackage[caption=false,font=normalsize,labelfont=sf,textfont=sf]{subfig}
\usepackage{booktabs}
\usepackage{natbib}

\title{EfficientECG: Cross-Attention with Feature Fusion for Efficient Electrocardiogram Classification
}

\author{
  Hanhui Deng, Xinglin Li, Jie Luo, Di Wu \\
  Key Laboratory for Embedded and Network Computing of Hunan Province \\
  Hunan University \\
  Changsha\\
  \texttt{\{denghanhui,lixinglin,luojie,dwu\}@hnu.edu.cn} \\
}

\begin{document}
\maketitle

\begin{abstract}
Electrocardiogram is a useful diagnostic signal that can detect cardiac abnormalities by measuring the electrical activity generated by the heart. Due to its rapid, non-invasive, and richly informative characteristics, ECG has many emerging applications. In this paper, we study novel deep learning technologies to effectively manage and analyse ECG data, with the aim of building a diagnostic model, accurately and quickly, that can substantially reduce the burden on medical workers. Unlike the existing ECG models that exhibit a high misdiagnosis rate, our deep learning approaches can automatically extract the features of ECG data through end-to-end training. Specifically, we first devise EfficientECG, an accurate and lightweight classification model for ECG analysis based on the existing EfficientNet model, which can effectively handle high-frequency long-sequence ECG data with various leading types. On top of that, we next propose a cross-attention-based feature fusion model of EfficientECG for analysing multi-lead ECG data with multiple features (e.g., gender and age). Our evaluations on representative ECG datasets validate the superiority of our model against state-of-the-art works in terms of high precision, multi-feature fusion, and lightweights.
\end{abstract}


\section{Introduction}
\label{sec:introduction}
Electrocardiogram (ECG) is one of the most important tools for detecting heart diseases in clinical diagnosis. It reflects the state of the heart when beating through a set of time-sequential sampling signals. It is a significant reference item for various cardiology diagnoses~\cite{He2024TII}. The patient cannot be adequately treated unless the type of arrhythmia is determined early and accurately. As a result, how to quickly classify the type of arrhythmia has been an important research topic. With the advancement of technology and the continuous evolution of detection methods, ECG data has gradually expanded from the previous single-lead to the 6-lead, 8-lead, and 12-lead currently~\cite{BhaskarpanditTIM2023}. The wide application of ECG and the increase in the number of leads have led to an increase in ECG data. In the meantime, some cardiology research has shown the impact of personal attributes such as age and gender features on the prevalence of heart diseases, which means additional patient data could potentially influence the pattern of ECG.~\cite{ArteagaTIM2016}. However, previous methods of manual diagnosis are inefficient in handling such a large volume of data.



Computer-aided auxiliary approaches have been playing increasingly important roles in ECG diagnosis.~\cite{Wang2025TII} Whereas the existing ECG interpretation algorithms still show a high misdiagnosis rate due to unsatisfactory detection and location of specific waves in a heartbeat and a short record~\cite{SHAH2007385}. Therefore, there is a pressing need to study how to use these massive data with manual labels to build an accurate and efficient model to help doctors perform auxiliary diagnosis. Benefitting from its powerful ability for data mining, machine learning research including deep learning has emerged in various data-driven areas in the last few decades~\cite{HeCVPR2016}. In this case, there have been some studies that investigated the application of machine learning in ECG diagnosis.

%
However, the existing studies on ECG classification suffer from two limitations:

\vspace{3pt} \noindent \textbf{Motivation 1.}
AI-supported accurate and rapid classification of arrhythmia diseases based on ECG data has long been a hot issue in academic research. In the early years, the most commonly used method for the ECG data classification model was machine learning, which builds the model through manual analysis and feature extraction. Previous work based on traditional machine learning has produced several achievements~\cite{Martis2013,Li2016ECGCU}. However, these methods usually require laborious manual feature engineering, which could be time-consuming and have problems with model generalisation ability to other types of data. In recent years, there has been much research focusing on deep learning for processing ECG data. Acharya \textit{et al.} \cite{Acharya2017} proposed the augmentation with CNN model. Kachuee \textit{et al.}~\cite{KachueeICHI2018} constructed an expanded dataset with 5 categories based on the collated MIT-BIH dataset, and built the classification model on the basis of ResNet~\cite{HeCVPR2016}. Hannun \textit{et al.}~\cite{Hannun2019} built an end-to-end deep neural network based on the ResNet network. However, due to the increasing number of layers in ResNet which leads to massive parameters, the computational resource consumption of this approach is prohibitively expensive, which hinders the application of similar deep learning models at real-time diagnosis~\cite{Qi2022TII}.

\vspace{3pt} \noindent \textbf{Motivation 2.}
Previous studies have shown that the deep learning model can obtain relatively satisfactory results in the utilization of ECG data. 
However, there are problems, such as too few samples and concerns about patients' privacy. As a result, most of the ECG research in past decades is based on the MIT-BIH dataset~\cite{MoodyIEMBM2001} which was first released in 1980. During these years, an increasing number of newly collected ECG datasets have emerged. In 2017, PhysioNet organised the Computing in Cardiology (CinC) Challenge~\cite{CinC2017}, which provided 8,528 single-lead ECG data samples, including five types of ECG events. However, the equipment used is still single-lead. There is still a big gap between the accuracy of used equipment and the multi-lead ECG equipment used by medical institutions, and the displayed ECG information is not as rich as that of multi-lead equipment. In 2019, to further discover the potential of AI algorithms in ECG data, Alibaba Cloud Computing hosted the 2019 ECG Human-Machine Intelligence Competition (HMIC), with a dataset of up to 40,000 8-lead ECG data~\cite{HMIC} provided by local hospital and university. In addition to ECG sampling data, it also contains the corresponding patient’s gender and age labels, paving the way for further research on the application of deep learning in the field of ECG based on a variety of feature types.

\vspace{3pt} \noindent \textbf{Our Approaches.} Motivated by the aforementioned problems, this work focuses on the classification and diagnosis tasks of high-frequency ECG data. Firstly, we discuss the sampling principle and data characteristics of ECG data, and compare the advantages and disadvantages of the previous studies and recent developments. Depending on the specific task, feature engineering was conducted. Then, by changing the data input head and adding an attention mechanism for multi-feature ECG analysis specifically, we transfer and optimise EfficientNet~\cite{TanICML2019}, one of the mainstream backbone models in the CV area. By inheriting the merits of various modules, we construct an efficient, accurate, and stable ECG classification structure, namely EfficientECG, to classify ECG data. Next, the high-quality, popular and public dataset in the field of ECG is used for evaluation. For the multi-feature dataset, including age and gender features, multi-feature fusion and joint training are carried out to further improve the classification performance of our proposed model on ECG.

\vspace{3pt} \noindent \textbf{Contributions.}
The major contributions of this work are listed as follows:
\begin{itemize}
  \item
 We investigate an ECG time sequence data classification model based on an EfficientNet neural network and multi-feature fusion. According to the characteristics of the ECG data and specific task, we design feature engineering and optimise the model structure based on the EfficientNet model, which is suitable for multi-lead ECG data.
   \item
   Based on the improved EfficientNet model suitable for ECG data, we further study the effect of the ECG data's external features on the ECG classification effect. For the additional external multi-features of age and gender, we devise a specific data preprocessing process and model construction that includes a cross-attention module to extract the multi-modal feature.
   \item
   We perform a thorough experimental evaluation. We use three authoritative datasets in the field of ECG to conduct evaluation, and compare the model structure proposed in this article with previous methods, showing the efficiency and effectiveness of our proposed EfficientECG model from both total parameters and accuracy metrics.
\end{itemize}

The remainder of this paper is organized as follows. Section~\ref{sect:related} summarizes related works on learning for ECG. Section~\ref{sect:classification} introduces our classification approaches in EfficientECG. Section~\ref{sect:fusion} describes our multi-feature fusion approaches in EfficientECG. Section~\ref{sect:experiments} presents experimental results. Section~\ref{sect:conclusion} concludes our paper.

\section{Related Work}
\label{sect:related}

\subsection{Traditional Machine Learning}
The most commonly used method in previous work on ECG analysis is machine learning. Early machine learning research generally focuses on feature extraction. Chazal and Reilly \cite{1202346} constructed 12 artificial features after a detailed analysis of ECG data. Based on MIT-BIH data, they constructed ECG data with 5 types of labels, and built a machine learning model. Martis \textit{et al.}~\cite{Martis2013} analysed the R wave in the ECG data, sampled near the waveform, constructed feature data, used a traditional machine learning model for classification, and used support vector machines (SVM) for classification on the MIT-BIH dataset. From the point of view of signal processing, Korurek \textit{et al.} \cite{Korurek2010} used time-domain feature extraction on sequence data and combined PCA for feature dimensionality reduction, then built a machine learning model. Totally, the ECG classification work on traditional machine learning has earned decent results. Nevertheless, traditional machine learning strongly relies on the features themselves and requires manual feature analysis and extraction. The model constructed in this way strongly depends on the accuracy of manual feature extraction and the importance of features~\cite{Nguyen2017}.

\subsection{Deep Neural Networks}
Neural networks often do not require tedious data preprocessing and feature engineering. This makes the neural network particularly well-suited to accomplish complicated tasks with vast amounts of data, even with manual data labelling. In recent years, with the improvement of neural network algorithm performance, research on ECG classification based on deep learning algorithms has also begun to emerge \cite{MaTIM2021}. Yu~\textit{et al.}~\cite{YuICPADS2018} researched the method of using edge devices to process health data based on CNN. Salloum and Kuo \cite{SalloumICASSP2017} proposed using RNNs to solve the classification of ECG data, because RNNs are well suited to processing sequential data \cite{Wu2019}, such as an ECG signal. In recent years, related research has begun to transfer to the application of model structures, which perform well in the field of computer vision, to improve the effect of ECG data. Kachuee \textit{et al.} \cite{KachueeICHI2018} used the ResNet model, which was widely used in the field of computer vision, to build the ECG classification model. Based on traditional ECG data classification work, Hannun \textit{et al.} \cite{Hannun2019} developed an end-to-end deep neural network (DNN) based on the ResNet network. Rohr \textit{et al.}~\cite{Rohr2022PM} presented ECG-RCLSTM-Net, consisting of a ResNet designed to analyse global features and a CNN-LSTM architecture (CLSTM) which analyses local features. As deep learning has proven to be a promising way to solve the ECG classification task on massive data, there are still aspects to improve, e.g., accuracy, and the number of parameters.

\section{EfficientECG: Model for ECG Classification}
\label{sect:classification}

We present the methods for our improved EfficientNet neural network, and show how to combine feature engineering with EfficientNet structure to build an efficient and accurate ECG classification model.


\subsection{Problem Formulation}
The procedure for the EfficientNet model to process ECG data and output the classification result is as follows: After data collection and storage, there is a dataset of $N$ single-lead ECG data samples. Each ECG data sample can be formalized as: $(x_{i1},x_{i2},...x_{i(T-1)},x_{iT},y^{(i)})$, in which $T$ denotes the number of time steps, corresponding to the amount of ECG data sampling points. $x$ represents the ECG data sequence. According to the sequence $x$, we build a neural network which can automatically perform feature extraction and train a classification model to generate the predicted value $\hat{y}^{(i)}$. The value $\hat{y}^{(i)}$ is associated with a specific kind of category label, and it can be calculated by using the softmax activation function to map the output of each category to a probability between $[0,1]$.

\subsection{Feature Engineering}
An effective feature engineering design requires both domain expertise and pattern recognition techniques according to existing data. 
As a case, in the PhysioNet 2017 dataset, the ECG data are related to atrial fibrillation (AF) signal detection and classification. Based on the previous study, AF detection methodologies fall into two main categories: those centered on analyzing atrial activity and those focused on assessing ventricular response~\cite{wijesurendra2019mechanisms}. Atrial activity analysis primarily involves scrutinizing the absence of P-waves or the presence of fibrillation F-waves in the TQ interval~\cite{purerfellner2014pwave}. Whereas ventricular response analysis methods concentrate on the beat interval of the QRS complex in the ECG, referred to as the RR interval~\cite{aeschbacher2018qrs}.

Through the background research mentioned above, this paper utilizes the bio-signal processing tool library BioSPPy~\cite{biosppy} for data preprocessing and feature engineering on the PhysioNet dataset. Considering the research on atrial fibrillation detection methods mentioned above as well as multiple rounds of feature combination validation, this study eventually selects the R-peak and P wave as constructed features for extra feature learning. According to the needs of batch training the model, the extracted R-peak and P-wave sequence data are clipped and padded into batches. Additionally, a masking layer is used in subsequent steps to eliminate the overfitting impact caused by padding values.

In the preprocessing and feature selection, we first conducted 3\textasciitilde45 Hz Finite Impulse Response (FIR) bandpass filtering on the original ECG signal to better reduce the impact of noise on the signal, and also standardize the ECG signal. Subsequently, R peak detection is conducted based on the Hamilton Segmenter algorithm, and P wave detection is further conducted based on the R-peak position. Considering that the located sequences of R-peaks and P-waves in the extracted ECG signal are relatively sparse and low-frequency compared to the original signal distribution, an AutoEncoder (AE) structure based on LSTM networks is used for feature extraction. This approach could effectively extract temporal features without making the model parameters relatively redundant.

\subsection{Model Architecture Design}
\subsubsection{Mobile inverted bottleneck (MBConv).}
\begin{figure}[tb]
\centering
\includegraphics[width=0.4\columnwidth]{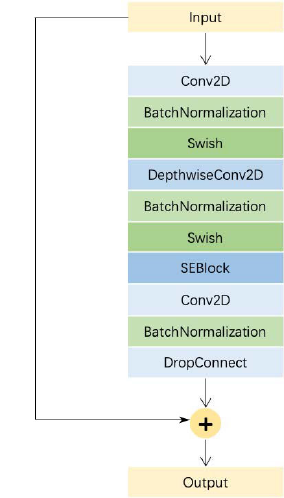}
\caption{The architecture of MBConv.}
\label{fig: mbconv}
\end{figure}

EfficientNet~\cite{TanICML2019} use mobile inverted bottleneck MBConv~\cite{SandlerCVPR2018} as its main building block. MBConv in our EfficientECG contains a squeeze-and-excitation block (SEBlock) and depthwise separable convolution (DSConv) layer, with basic layers such as BatchNormalization and Conv2D to build its basic layers. The structure of MBConv in this work is shown in Figure~\ref{fig: mbconv}. In detail, DSConv \cite{CholletCVPR2017} replaces the full convolutional operator with two separate layers, depthwise convolution and pointwise convolution. Compared to standard full convolution, MBConv employs several smaller filters with fewer parameters. Along with the reduction of computation costs, the risk of overfitting can be mitigated. SEBlock \cite{HuCVPR2018} has two main operators, which are squeeze and excitation. The squeeze operator has the ability to shrink feature maps across spatial dimensions. Excitation can learn a parameter $W$ to produce channel-wise weights. According to the weight of each feature channel, we can selectively enhance useful features, suppress less useful features, and consequently improve network performance. As a result, MBConv not only has the merit of fewer parameters and less computation cost from DSConv, but also gains the effect of SEBlock's multi-scale feature fusion. 

\begin{figure*}[tb]
\centering
\includegraphics[width=0.8\columnwidth]{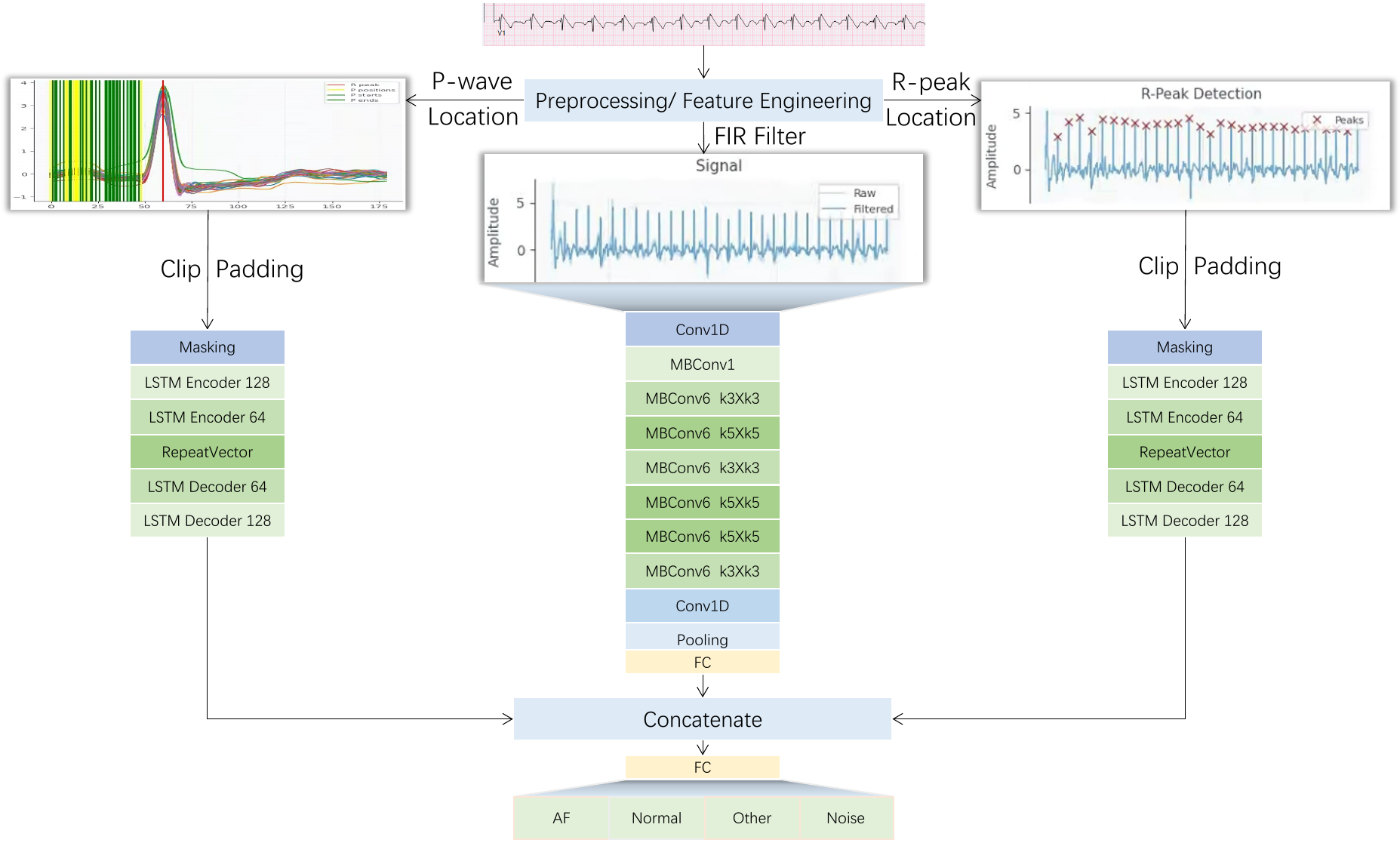}
\caption{The architecture of EfficientECG. The raw ECG records are firstly preprocessed by bio-signal feature engineering, the obtained P-wave and R-peak feature are extracted by LSTM AutoEncoder, and filtered ECG signals are passed to a customized EfficientNet, the three extracted features are combined as the input of classification.}
\label{fig:efficientecg}
\end{figure*}

\subsubsection{Optimization of model structure for ECG data.}
To meet the long sequence processing and high real-time requirements of ECG data, our EfficientECG has made improvements according to the ECG sequence data based on the EfficientNet network structure, so that it can obtain good results on ECG data. Two improvements are described below.

The first improvement is for the feature extraction module of ECG data. Based on the ECG data structure, we rationally adjust the convolution dimension in the feature extraction module. In the process of data collection and storage, ECG data will be stored in a Numpy matrix of size $C*N$, where $N$ represents the length of the time step, which is the number of sampling points, and $C$ represents the sampling channel, corresponding to the lead number of the ECG data. For example, the shape of single-lead ECG data is $1*N$. Image data in EfficientNet \cite{TanICML2019} is often composed of a three-dimensional matrix with width and height and the number of channels, while ECG data is composed of a two-dimensional matrix with the number of channels and length, and there is a difference in time order. Directly applying EfficientNet's $3*3$ convolution module is not suitable. Based on the differences in the above data structure, we converted the original EfficientNet's $3*3$ Conv module to the Conv1D structure so that feature extraction can be performed using multiple $1*N$ sliding filters. In this case, feature extraction is to perform convolution extraction along the axis of the time sequence, which can extract the features of partial signal fluctuations in the ECG data.

Secondly, the ECG data sampling frequency is extremely high, usually 500 $Hz$. If the sampling duration is 10 $s$, a long sequence of 1*15,000 will be generated. When the features of the data are extracted through Conv1D and several MBConv layers, a very long set of feature tensors will be generated. In this case, if the original structure of a single fully connected layer is still used, it will not be enough to completely aggregate all the extracted features, which leads to falling results. To adapt to the characteristics of ECG data, we added one more fully connected layer. To reduce the impact of overfitting caused by the addition of the fully connected layer, we employ the dropout method in the fully connected layer, and use $L_2$ regularisation for the fully connected layer. It improves features' aggregation ability and further prevents the risk of overfitting.

In summary, through the above description of the model structure, we can use the MBConv layer structure as the main building block of the model based on the DSConv and SEBlock to build our personalised ECG classification model and make some improvements for the characteristics of high-frequency ECG time sequence data. 

The architecture overview of our EfficientECG is shown in Figure~\ref{fig:efficientecg}, where we can see that EEG data is firstly processed through feature engineering including ECG signal preprocessing. The filtered ECG signal is through the Conv1D layer for feature conversion, then further extracted and enhanced through multiple layers of MBConv layers. Subsequently, feature filtering is performed through the pooling layer. Meanwhile, the R-peak and P-wave localization features extracted based on the original ECG signal are subjected to feature extraction through an LSTM autoencoder. These three types of features are then fused through the Concatenate layer. Finally, Softmax probability mapping is performed through the fully-connected layer.

\subsection{Model Training}
The EfficientECG network structure proposed in this paper is used for model training based on the above improvements. Given single-lead ECG data samples formalized as: $(x_{i1},x_{i2},...x_{i(T-1)},x_{iT},y^{(i)})$, by modeling the sequence $x$, a classification model is trained to generate the predicted value $\hat{y}^{(i)}$. The mean squared error loss function and the $L_2$ regularisation are used as the loss function of the model, using the true value $y^{(i)}$, the predicted value $\hat{y}^{(i)}$ and the weight matrix of every layer ${W}^{[l]}$ to compute the loss. The loss computation formula is as follows:

\begin{equation}
\centering
loss(\hat{y},y)=\frac{1}{m} \sum\limits_{i=1}^{m} (y^{(i)}-\hat{y}^{(i)})^2 + \frac{\lambda}{m}\sum\limits_{l=1}^{L} {\|{W}^{[l]}\|}_{F}^2
\label{equ1}
\end{equation}
After the loss computation is completed, the gradient descent algorithm and the backpropagation algorithm are used to train the model, calculate the gradient, adjust the model weight, and gradually approach the optimal value. In our EfficientECG, the number of global iterations is used to control the learning rate variation. By using a small set of samples as the validation set to compute the gradient loss together during gradient descent, the loss can be accurately estimated and the anti-interference ability against outliers can be enhanced.

A fixed learning rate during the gradient descent would lead to instability of model convergence. In this case, for our EfficientECG, we adopt a learning rate variation method proposed by Vaswani \textit{et al.} \cite{VaswaniNIPS2017}, in which the learning rate varies with global iteration steps. The specific formula is as follows:

\begin{equation}
\centering 
\begin{split}lrate=&d^{-0.5}_{model}*\min({step\_num}^{-0.5},\\
&{step\_num}*{warmup\_steps}^{-1.5})
\end{split}
\label{equ2}
\end{equation}
where $lrate$ represents the learning rate, $warmup\_steps$ represents the number of slow training steps in the early stage, and $step\_num$ represents the number of steps in the current iteration. According to the formula above, in the early stage of model training, the learning rate will have a rapid rising period, and then as the number of iterations (step\_num) increases, the learning rate will gradually decrease and converge to a smaller value. This setting is consistent with the model's convergence process: the parameter adjustment range is large at the early stage, converges at the latter stage, and tunes locally. Through this method, the model can further fit the training data and converge to a better result.

To avoid overfitting, EfficientECG utilizes the EarlyStopping module in the Keras deep learning framework, which can specify the number of validation rounds, the number of validation data, and the monitoring of the evaluation metrics, stopping training in time. By using early stopping, the model can converge to a certain level before overfitting, then stop the training in time, and save the current optimal model weight.

\section{EfficientECG for Multi-feature Fusion}
\label{sect:fusion}

In this section, we further verify the performance of the EfficientECG model proposed on multi-lead ECG data with multi-feature analysis. Through data analysis on the HMIC dataset, we noticed that age and gender are also important features that affect classification. Based on the basic architecture of EfficientECG, the model is improved to make it suitable for multi-feature input. Gender, age, and ECG data characteristics are fused to build an end-to-end model.

\subsection{Data Analysis on Multi-Feature ECG Data}
\begin{figure}[htb]
\begin{center}
\includegraphics[width=0.4\columnwidth]{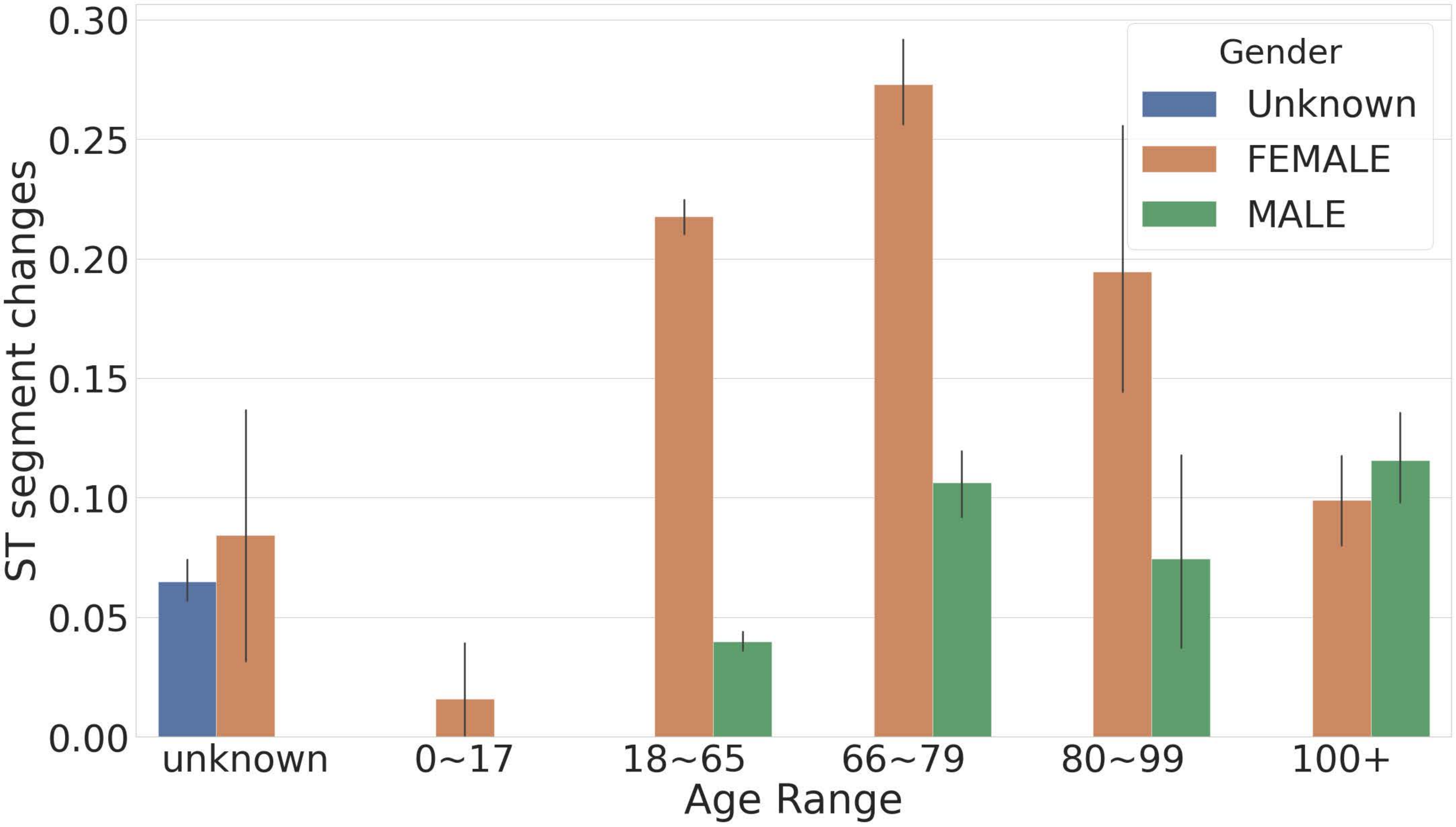}\\ 
{\scriptsize(a) The distribution of ST segment changes.}\\
\includegraphics[width=0.4\columnwidth]{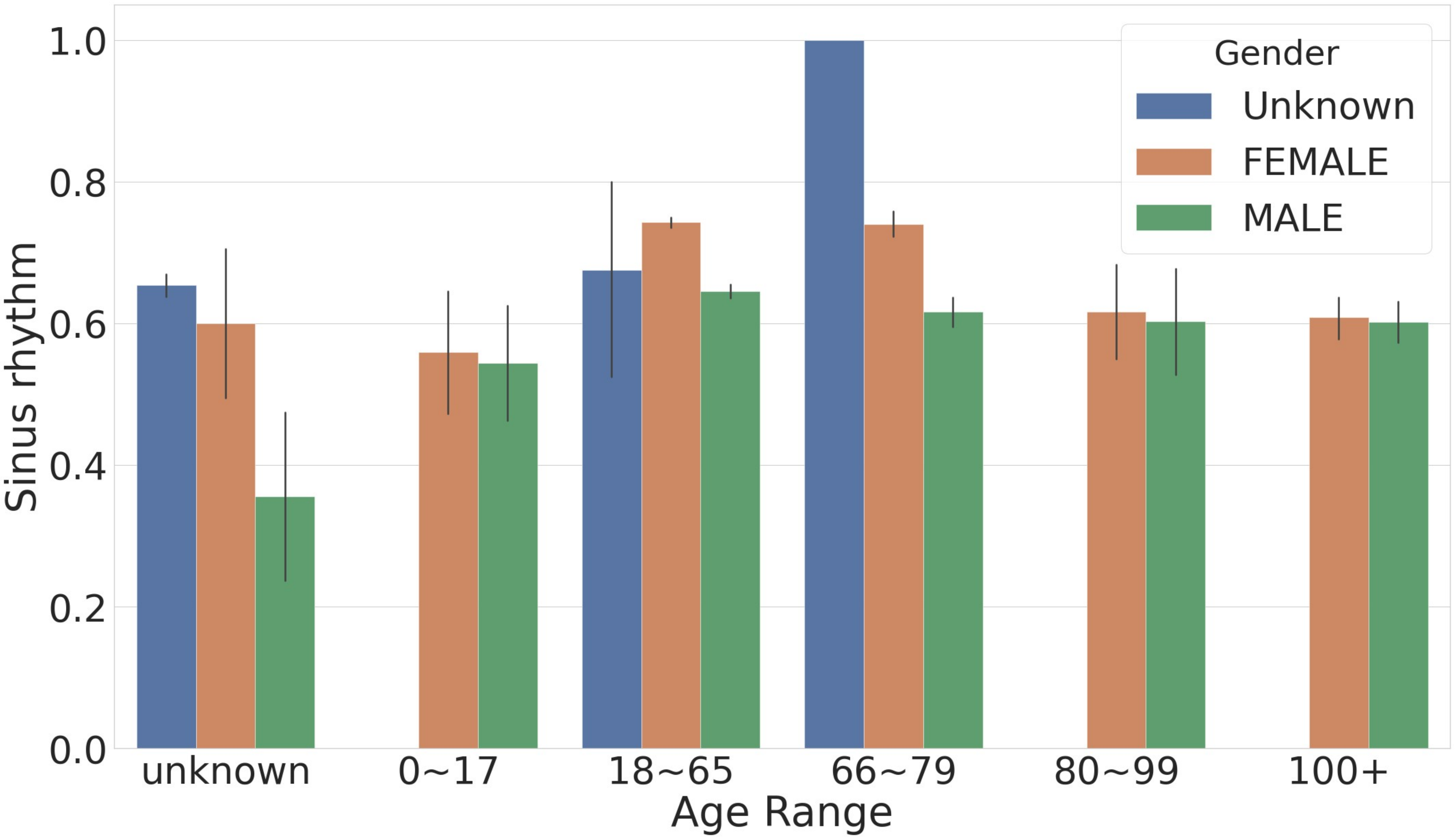}\\
{\scriptsize (b) The distribution of sinus rhythm}
\caption{The distribution of two arrhythmias in various ages and genders.}
\label{fig:analy}
\end{center}
\end{figure}

We conducted a detailed analysis of the existing 2019 HMIC dataset, in which 32,142 ECG cases were horizontally compared according to different ages and genders. It was found that some types of heart diseases are different in their distribution among a variety of ages and genders. Figure~\ref{fig:analy} depicts one of the results, which shows that there are significant differences in the distribution of ST segment changes across ages and genders. However, the distribution of Sinus rhythm does not obviously vary with age and gender features, meaning that age and gender features can help classify different ECG categories with statistical significance.

\subsection{Multi-Feature Module Design}
Multi-feature means external features, including the age and gender of patients, combined with the ECG data to perform feature extraction and feature fusion, which leads to the building of the classification model. There are two points. The first is to figure out how to extract the feature of different ages so that the age of similar ECG expressions has a similar vector representation. In other words, how can the model learn the relationships between different ages and genders automatically? The other point is how to construct an end-to-end joint model for training after multi-scale feature fusion.

\begin{figure}[htb]
\centering
\includegraphics[width=0.8\columnwidth]{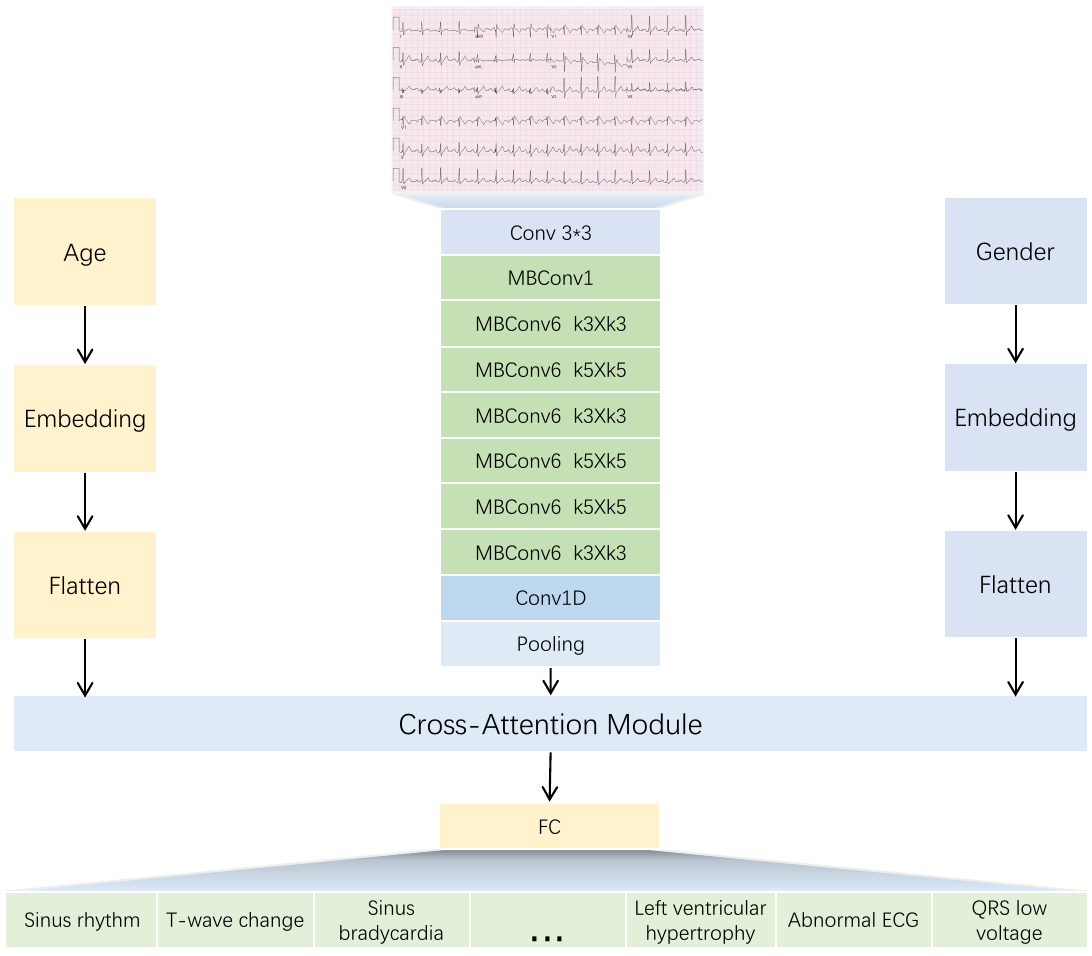}
\caption{The architecture of multi-feature fusion in EfficientECG.}
\label{fig:multi-feature}
\end{figure}

Based on the EfficientECG model proposed in the previous section to improve classification, we present the extraction module of gender and age features, as well as the cross-attention module, to jointly construct a multi-feature fusion classification model. The specific structure is shown in Figure~\ref{fig:multi-feature}. Age and gender features are added separately. After the Embedding layer expands the feature dimensions, the dimensional conversion is performed through the Flatten layer. Gender and age are aggregated with ECG data features via a cross-attention module, and finally, fully connected layers are used for probability mapping. In addition, the label of the data is a multi-label of 55 categories, and there is no mutual exclusion relationship between categories. Thus, the activation function used on the last fully connected layer needs to be modified to a sigmoid for probability mapping.

\subsection{Feature Extraction based on Embedding Layer}
For learning the hidden relationship between input features, and obtaining the representation of high-dimensional features, the common method is to construct an embedding layer, which can perform feature conversion to convert the input numerical features into a higher-dimensional vector representation. Through high-dimensional vectors, the hidden relationships between features can be learned. Based on this idea, we apply it to the external features of ECG cases to learn the hidden representation of external features. We use the embedding layer to transform the dimension of the age and gender features in our input data, and expand the original 1-dimensional input features into a $1*D$ dimensional feature representation, where $D$ represents the output vector dimension of the Embedding layer. Through the iterative training of the model, it will automatically adjust the weight of the embedding layer, discover the connection between the vectors, and obtain a better representation of hidden features.

\subsection{Cross-Attention Module for Multi-Feature Fusion}

\begin{figure}[tb]
\centering
\includegraphics[width=0.8\columnwidth]{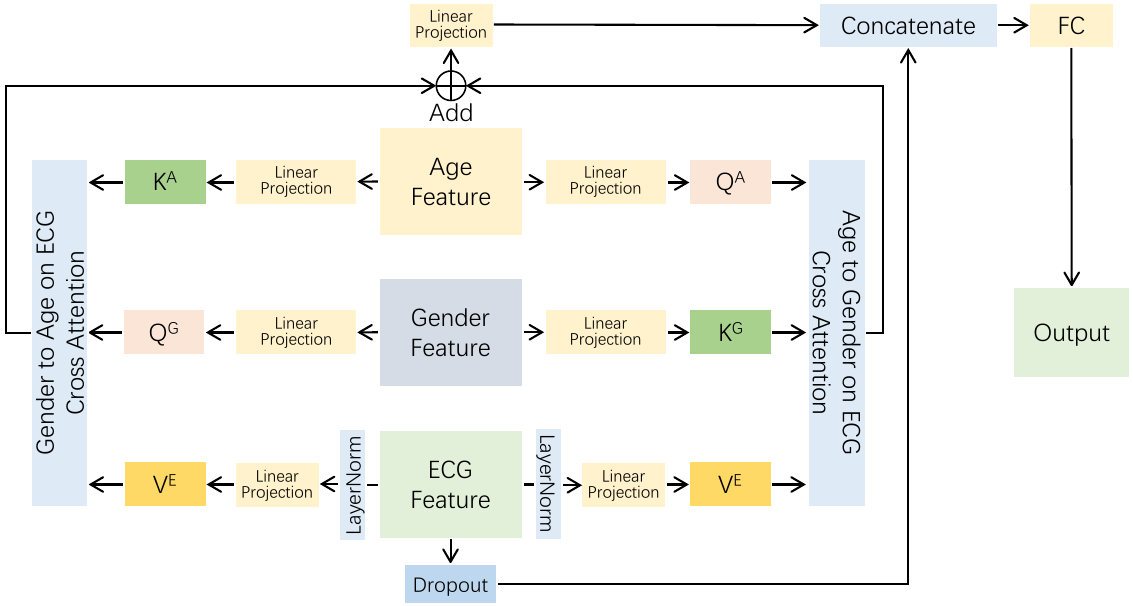}
\caption{The architecture of cross-attention module for multi-feature.}
\label{fig:cross-attention}
\end{figure}

Inspired by the success of the cross-attention mechanism in the computer vision area, we propose a multi-feature cross-attention module for the HMIC datasets of multi-feature. Figure~\ref{fig:cross-attention} illustrates the architecture of our cross-attention module. As shown in Figure~\ref{fig:multi-feature}, the input features involved in this module include the ECG feature extracted by the customised EfficientNet model, age and gender feature extracted by the embedding layer. To align the features of different shapes, the linear layer is utilized to project the features into the feature space, which means features share the same length of the last dimension. Benefiting from the feature alignment, the different features could be used for cross-attention. In our module, there are two cross-attention blocks which are age-to-gender and gender-to-age attention. Both attention blocks use the ECG feature as value tensor $V^E$. Age-to-gender attention uses the age feature as query tensor $Q^A$, the gender feature as key tensor $K^G$, and vice versa. The outputs of two attention blocks are added as the cross-attention output, which would be aligned again with ECG features via a linear layer subsequently. Then the cross-attention output and ECG feature are sent to a concatenate layer to build the fusion feature. This fusion feature would be through a fully connected layer with Sigmoid activation to obtain the classification results. The detailed calculation inside two cross-attention blocks can be represented as follows:
\begin{equation}
    \centering
    \boldsymbol{CA}=
     \begin{cases}
       \operatorname{softmax}\left(Q^A \times K^{G^{\mathrm{T}}}\right), Age \ to\  Gender;\\
       \operatorname{softmax}\left(Q^G \times K^{A^{\mathrm{T}}}\right), Gender \ to\ Age;
     \end{cases}
\end{equation}
\begin{equation}
  \centering
     \boldsymbol{Output}=\boldsymbol{CA} \times V^E
\end{equation}
The first equation is the calculation of cross-attention between age and gender features, where the projected feature is multiplied by the transposed other feature, and then goes through Softmax activation. In the second equation, the cross-attention calculated would be multiplied by the projected ECG feature to obtain the output.

\section{Evaluations}
\label{sect:experiments}

\subsection{Evaluation on Single-Lead ECG Data}
\subsubsection{Datasets and preprocessing}
To validate the results, the evaluation in this section uses two authoritative datasets. The first is the MIT-BIH dataset, whose original version was taken from~\cite{MoodyIEMBM2001}, and has later been recompiled by Kachuee \textit{et al.}\cite{KachueeICHI2018}, where over 4,000 samples were resampled into a single-label, five-category dataset with 109,446 ECG data points on only lead II, for the QRS complexes are not prominent on the lower signal of the other lead according to the official introduction of dataset. The sampling frequency was 125 $Hz$. We use the division and evaluation method of the dataset when referring to his work. The second is the PhysioNet dataset for the 2017 CinC competition. For the ECG data in the PhysioNet dataset, the data sampling duration is within 30s, and the data is divided into four classes with labels.

All datasets will go through preprocessing, which will include missing value processing and noisy value processing. For single-lead ECG data, if the ECG data is abnormal, the sample will be removed. The noise removal is not done too much, but combined with the strong anti-noise ability of the deep learning model, according to the type of sample and the number of samples, the data of each type of sample are expanded to enhance the generalisation ability of the model, further reducing the interference and influence of noise on the training process.

\subsubsection{Evaluation setup}
On the recompiled MIT-BIH dataset, we compare our EfficientECG model with the improved ResNet model \cite{KachueeICHI2018}, the DWT random forest model constructed by Li \textit{et al.} \cite{Li2016ECGCU}, the DWT SVM model constructed by Martis \textit{et al.} \cite{Martis2013}, and the Augmentation CNN model constructed by Acharya \textit{et al.} \cite{Acharya2017}, in terms of accuracy. For the ResNet model proposed by Kachuee, we further compared them using the precision and recall rate.

On the PhysioNet dataset, we compare our EfficientECG with the ECG-ResNet model proposed by Hannun \textit{et al.} \cite{Hannun2019}, ECG-RCLSTM-Net proposed by Rohr \textit{et al.} \cite{Rohr2022PM}, and MINA proposed by Hong \textit{et al.} \cite{HongIJCAI2019}, in terms of parameter amounts, average accuracy, and the CinC score based on F1 score. The ROC curve of each category using EfficientECG is drawn, and the AUC area is calculated. These two parameters are critical for judging the effect of a classifier in the current classification.

We used an 8:1:1 split for both datasets, with the training set accounting for 80\%, the validation set accounting for 10\%, and the test set accounting for 10\%. For the training process, we use Adam as the optimizer~\cite{Kingma2014} to train each model. According to the experience of model training, the maximum number of iterations is set to 100, and the threshold for early stopping of training is set for monitoring the loss changes of the model on the validation and training sets.

\begin{figure}[tb]
\begin{center}
\begin{tabular}{cc}
\includegraphics[width=0.4\columnwidth]{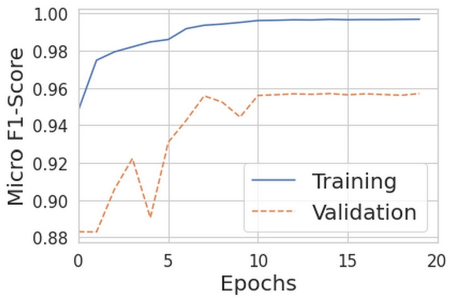} &
\includegraphics[width=0.4\columnwidth]{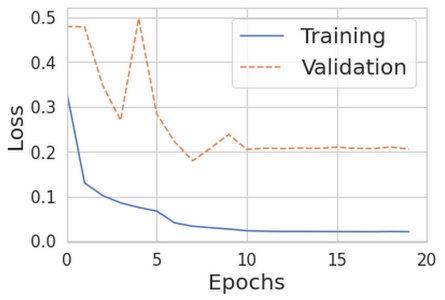}\\
{\scriptsize(a) Micro F1-Score} & {\scriptsize (b) Loss}
\end{tabular}
\caption{The accuracy and loss change of EfficientECG on MIT-BIH dataset.}
\label{fig:al-mit}
\end{center}
\end{figure}

The evaluation metrics used in this work are as follows:
\begin{itemize}
\item \textbf{F1 score.} The weighted average of precision and recall. It reaches the best value at 1 and the worst value at 0. The evaluation score used in the 2017 PhysioNet Competition is an improved version of F1. The weighted score is calculated based on the number of predicted tags. The final CinC score is calculated by averaging the F1 values of each category.
\item \textbf{Confusion matrix (CM).} Specific table layout that allows visualization of the performance of the classification effect. In this work, the X-axis represents the category of the predicted label, and the $Y$-axis represents the category of the true label. The diagonal line represents the number of correctly classified categories, and the remaining values represent the number of samples that were misclassified into other categories.
\item \textbf{Receiver Operating Characteristic (ROC) curve and Area Under the Curve (AUC).} The ROC curve usually has a true positive rate on the $Y$-axis and a false positive rate on the $X$-axis, and AUC is usually used for comparing the effects of the models. Generally, for a good model, the AUC would approach 1.
\end{itemize}

\begin{figure}[tb]
\begin{center}
\begin{tabular}{cc}
\includegraphics[width=0.4\columnwidth]{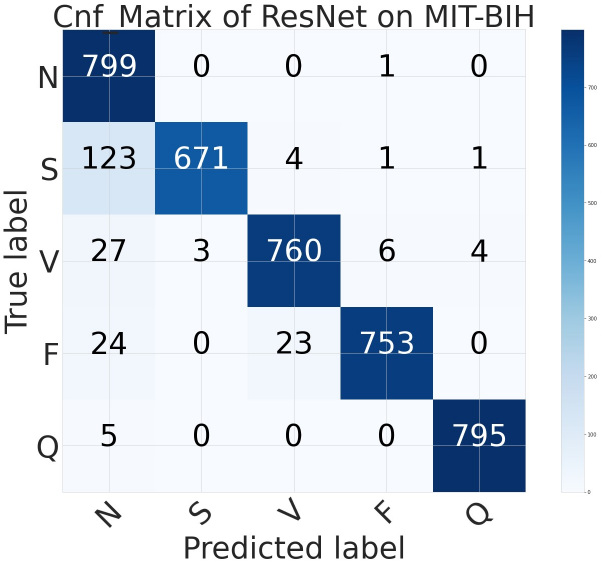} &
\includegraphics[width=0.4\columnwidth]{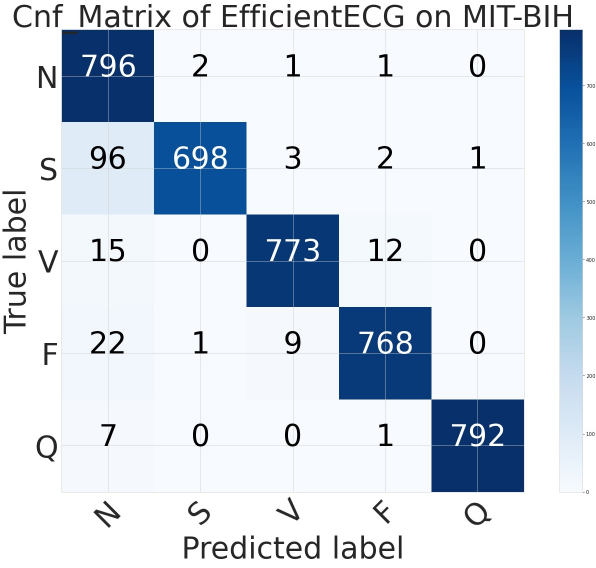}\\
{\scriptsize(a) CM of ResNet} & {\scriptsize (b) CM of EfficientECG}
\end{tabular}
\caption{CM of ResNet and EfficientECG on MIT-BIH dataset.}
\label{fig:cm-mit}
\end{center}
\end{figure}

\begin{table}[tb]
\centering \small \def\arraystretch{0.8}
\caption{The classification result on MIT-BIH dataset}
\begin{tabular}{llll}
\toprule
Work  &  Method  &  Avg acc &  F1\\
\midrule
\textbf{EfficientECG}     &  EfficientNet     &  \textbf{96.5\% }  & \textbf{0.96}\\
Kachuee \textit{et al.} \cite{KachueeICHI2018}  &  ResNet               &  93.4\%  & 0.87 \\  
Acharya \textit{et al.} \cite{Acharya2017}  &  Augmentation+CNN     &  93.5\%  & 0.83\\
Martis \textit{et al.} \cite{Martis2013}   &  DWT+SVM              &  93.8\%  & 0.78\\
Li \textit{et al.} \cite{Li2016ECGCU}       &  DWT+random forest    &  94.6\%  & 0.81\\
\bottomrule
\end{tabular}
\label{tab:mit}
\end{table}

\subsubsection{Results}
For the MIT-BIH dataset, we performed the evaluation procedures mentioned above. The change of accuracy rate during the model training process is shown in Figure~\ref{fig:al-mit}. It can be seen that the model has a small range of fluctuations in the iteration of about 3-4 epochs. Then it tends to be flat around the 10th step and converges to a better value.  The difference between the scores of the training set and the test set in the figure is due to the imbalance of the sample categories of the training set. By contrast, we selected 800 samples from each category to build the test set. Thus, the test set is a balanced sample set. With continuous training of the model, the loss converges. After about 20 epochs of training, when the early stopping threshold value we set in advance is reached, the training is automatically stopped, and the optimal model is saved. We use cross-entropy to compute the loss between the predicted value and the real value. The loss change is shown in Figure~\ref{fig:al-mit}. The model has a small range of fluctuations after iterating for 3-4 epochs, due to the same issue as above. It then tends to converge and obtain a superior classification around the 10th epoch. The comparison between ECG-Efficient and ResNet \cite{KachueeICHI2018} is shown in Figure~\ref{fig:cm-mit} in the form of confusion matrices.

\begin{figure}[tb]
\centering
\includegraphics[width=0.9\columnwidth]{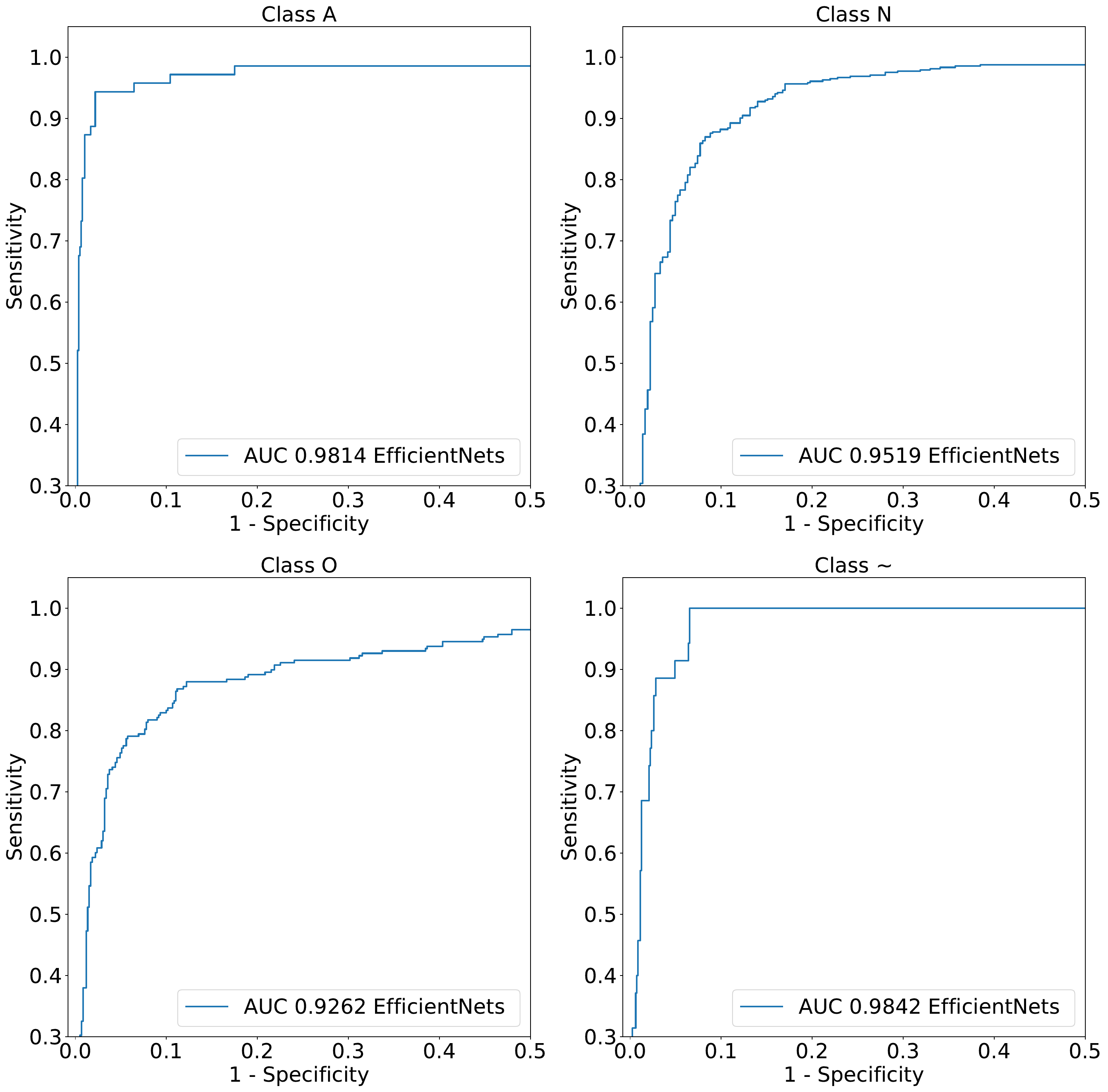}
\caption{ROC curve of 4 classes on PhysioNet dataset.}
\label{fig;roc-phy}
\end{figure}

\begin{table}[tb]
\caption{The classification result on 2017 PhysioNet dataset}
\centering \setlength{\tabcolsep}{3pt}
\scalebox{1}{$
\begin{tabular}{lrcccc}
\toprule
Model                         &  Total Para.  &  Avg Acc  & CinC \\
\midrule
\textbf{EfficientECG}             &  4,908,386      & \textbf{87.82}\%    &\textbf{0.862}\\
ECG-ResNet \cite{Hannun2019} &  10,473,892        & 86.90\%    &0.858\\ 
ECG-RCLSTM-Net \cite{Rohr2022PM} &  1,840,210        & 85.27\%   &0.805\\
MINA \cite{HongIJCAI2019} &  4,749,216        & 85.97\%    &0.850\\ 

\bottomrule
\end{tabular}$}
\label{tab:physionet}
\end{table} 

The detailed results are shown in Table~\ref{tab:mit}. The EfficientECG proposed in this section has a classification average accuracy (avg acc) of 96.5\% on the full test set, which is higher than ResNet \cite{KachueeICHI2018}, Augmentation CNN \cite{Acharya2017}, DWT SVM \cite{Martis2013} and DWT random forest \cite{Li2016ECGCU}. We made a more detailed comparison of the macro-F1 value of the ResNet model with EfficientECG. EfficientECG outperformed ResNet's 0.87 with a score of 0.96, which increases nearly 10\%. The evaluation results confirm that EfficientECG has superior feature extraction and aggregation capabilities in the ECG data classification task, and performs well in indistinguishable categories.

In addition, this paper retrains the proposed EfficientECG model based on the 2017 PhysioNet dataset. Figure~\ref{fig;roc-phy} is the ROC curve for each category of the EfficientECG model on the PhysioNet dataset. It can be seen from the figure that in the four ECG event categories, overall, the curve is close to the ideal point on the upper left and performs best in certain categories. The detailed evaluation results are shown in Table~\ref{tab:physionet}. By comparing with ECG-ResNet~\cite{Hannun2019}, ECG-RCLSTM-Net \cite{Rohr2022PM}, and MINA \cite{HongIJCAI2019}, we can see that generally, our EfficientECG own the best results for the both metrics. On the other hand, the total number of parameters used by EfficientECG is only 46\% of ECG-ResNet with similar classification results. In this case, when the model is training and predicting, it takes up less computing resources, and provides a better real-time response. Whereas ECG-RCLSTM-Net has fewer parameters than the above models, its classification accuracy and CinC score are relatively worse among the baseline models.

\begin{table*}[t]
\centering \small \def\arraystretch{0.4}
\caption{Evaluation on HMIC dataset,Micro-F1a on training set, Micro-F1b on test set}
\label{tab:hmic} 
\scalebox{1}{
\begin{tabular}{lrclll}
\toprule
Model                           &  Total parameters  & Batch size &  Dataset Features   & Micro-F1a &Micro-F1b  \\
\midrule
\textbf{EfficientECG}               &  4,121,459         & 64           &  8-lead              & 0.8506             &\textbf{0.8343}\\

Transformer\cite{VaswaniNIPS2017}&  250,728,031       & 16           &  8-lead              & 0.8211                &0.4100\\ 
ECG-ResNet\cite{Hannun2019}  &  11,139,031        & 32           &  8-lead              & 0.8315                &0.8230\\ 
EfficientNet\cite{NonakaCinC2020}                &  13,680,243         & 2           &  8-lead             & 0.8410                &0.8191\\
MINA\cite{HongIJCAI2019}               &  4,404,631         & 128           &  8-lead              & \textbf{0.8513}               &0.7996\\
ResNet-DenseNet\cite{HongIJCAI2019}               &  4,507,191         & 64           &  8-lead              & 0.6414             &0.6405\\
\midrule 
\textbf{EfficientECG}  &  4,707,317         & 8       &  8-lead+age+gender   & 0.8772                &\textbf{0.8661}\\ 
W/o gender data  &  4,532,244         & 8       &  8-lead+age  & \textbf{0.9206}                &0.8637\\ 
W/o age data &  4,532,244         & 8       &  8-lead+gender  &  0.8743                &0.8607\\ 
W/o cross-attention  &  4,142,195         & 8       &  8-lead+age+gender  & 0.8754                &0.8560\\ 
\bottomrule
\end{tabular}
}
\end{table*}

\begin{figure}[htb]
\begin{center}
\begin{tabular}{cc}
\includegraphics[width=0.4\columnwidth]{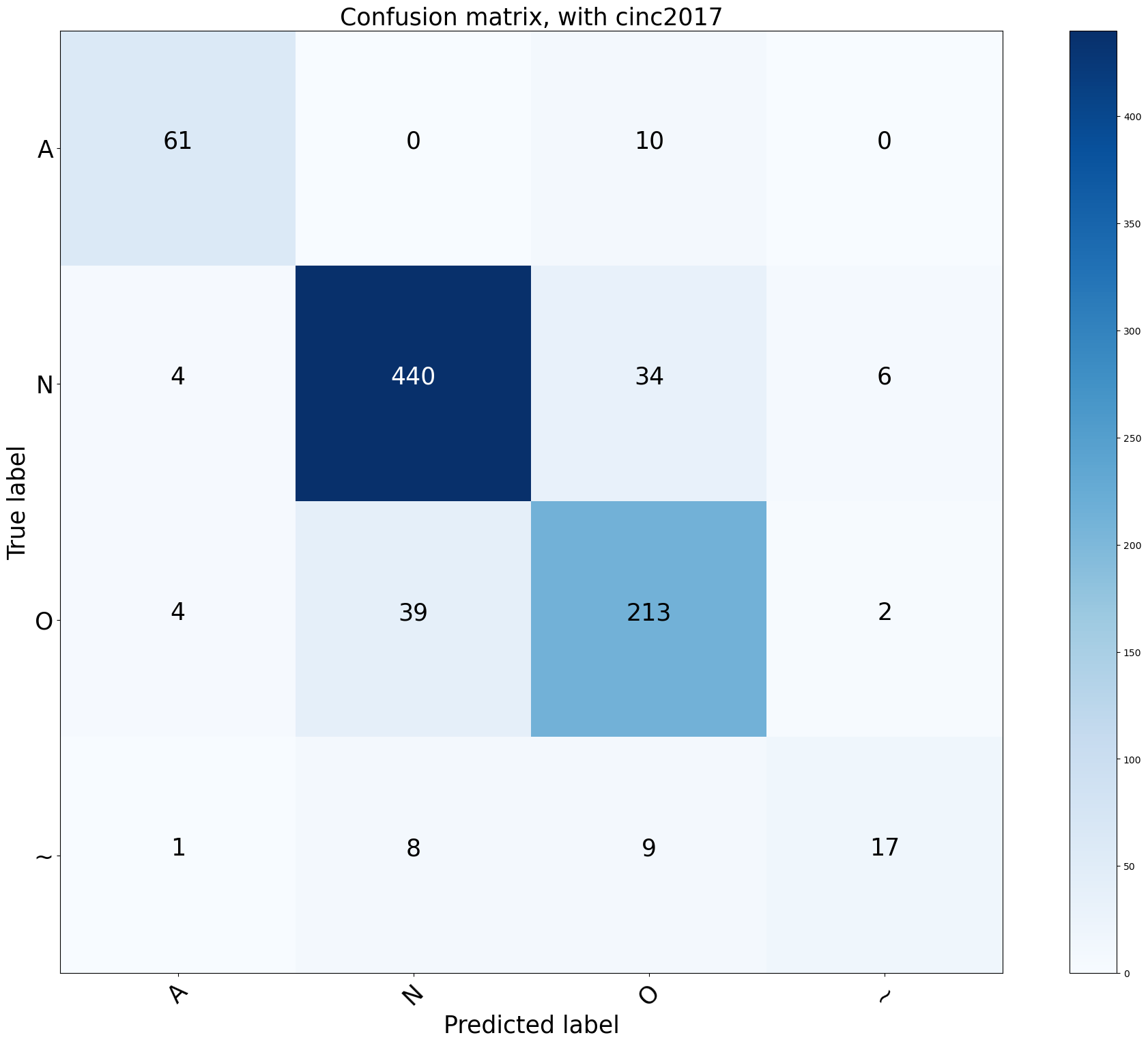} &
\includegraphics[width=0.4\columnwidth]{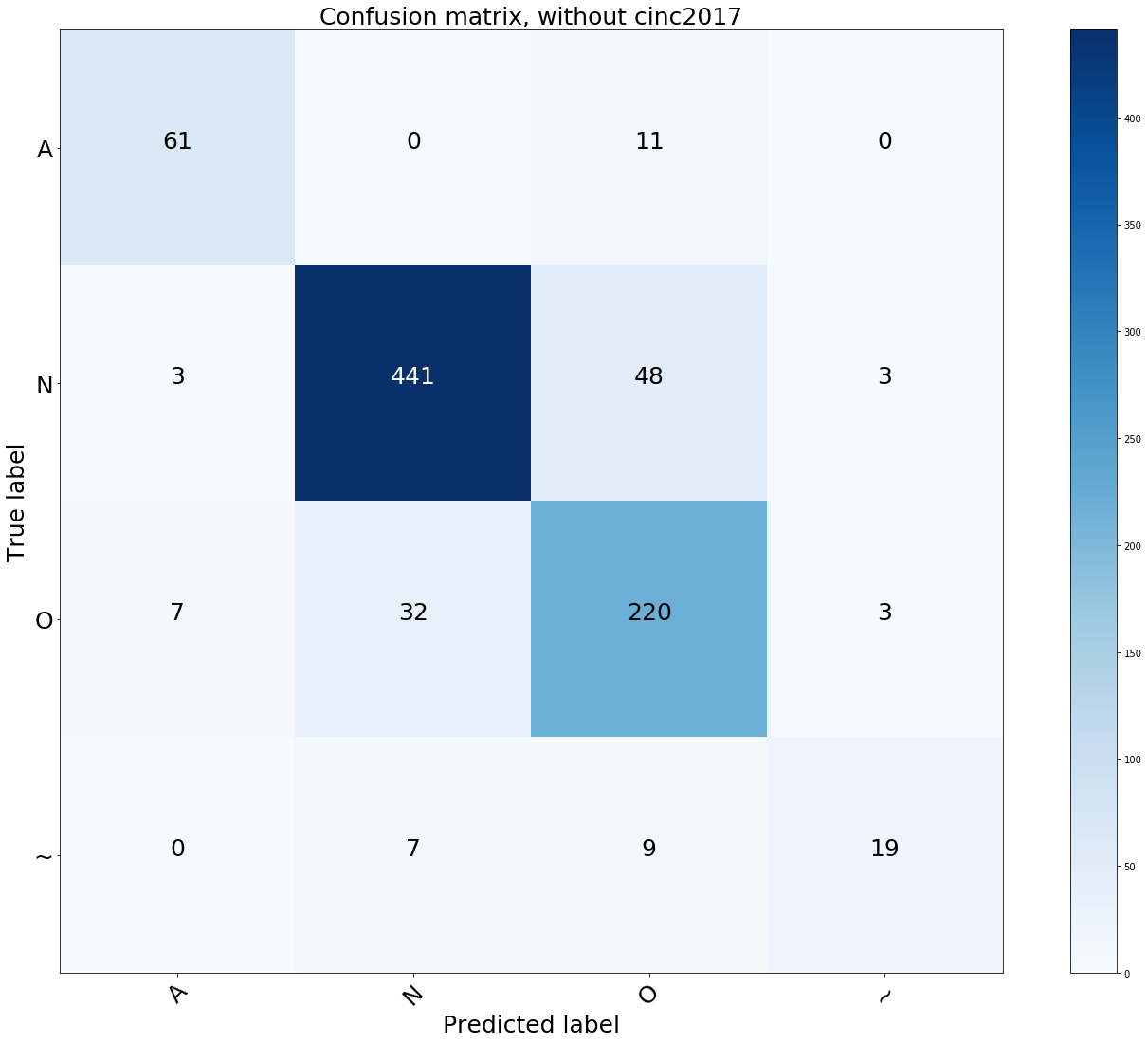}\\
{\scriptsize(a) With bio-signal} & {\scriptsize (b) Without bio-signal}
\end{tabular}
\caption{CM on PhysioNet dataset using bio-signal analysis}
\label{fig:cm-cinc}
\end{center}
\end{figure}

Particularly, regarding the improvement of specific classification tasks, we employed bio-signal analysis for feature engineering as mentioned previously. We carried out targeted feature selection based on the purpose of classification of atrial fibrillation. The specific classification effects before and after using this feature engineering are compared, as shown in Figure~\ref{fig:cm-cinc}. In terms of the overall CinC score, after applying bio-signal analysis for feature engineering, the score increased from 85.8 to 86.2, with a modest improvement. However, in classification A, which corresponds to the most important classification task of atrial fibrillation, the Micro-F1 value increased significantly from 0.847 to 0.865, indicating the effectiveness of the model in actual atrial fibrillation detection tasks. This also further demonstrates the importance of combining and analyzing domain-specific knowledge with deep learning in specific applications.

\subsection{Evaluation on Multi-Feature Data}
\subsubsection{Dataset preprocessing}
In this part, we will use a multi-feature dataset of the ``2019 Hefei High-tech Cup'' ECG Human-Machine Intelligence Competition (HMIC)~\cite{HMIC} to verify the improvements of EfficientECG on multi-lead data processing and feature fusion. Each sample in the HMIC dataset has 8-lead ECG signals associated with age and gender and age label. The sampling frequency is 500 $Hz$, and the duration is 10$s$. The dataset used here is public and contains 32,142 samples in total.

\subsubsection{Evaluation setup}
The EfficientECG is again compared with the ECG-ResNet \cite{Hannun2019} proposed by Hannun \textit{et al.} and MINA proposed by Hong \textit{et al.}~\cite{HongIJCAI2019}. In addition, we add the Transformer model proposed by Vaswani \textit{et al.} \cite{VaswaniNIPS2017} into comparison, as well as another EfficientNet-based ECG classification model proposed by Nonaka \textit{et al.} \cite{NonakaCinC2020}, and ResNet-DenseNet proposed by Hwang~\cite{hwang2023ResDense}. Since the dataset in this part is multi-lead, the input and output layers of the above models have been modified. Moreover, this dataset is multi-label data with 55 categories. Unlike the data in the previous section, there is no mutual exclusion between the labels. Therefore, when the final output is computed, a threshold must be defined for each category to obtain the probability. The calculation of the loss function must also be modified to use multi-label rather than multi-class classification. In this case, we use the cross-entropy loss, as it converges faster on multi-label tasks than MSE. The specific loss formula is as follows:
\begin{equation}
\centering
\begin{split} \textstyle
L=-\frac{1}{N}\sum_{i}[y^{(i)}\cdot \log(p_{i})+(1-y^{(i)})\cdot \log(1-p_{i})]
\end{split}
\label{equ3}
\end{equation}

\begin{figure}[tb]
\begin{center}
\begin{tabular}{cc}
\includegraphics[width=0.4\columnwidth]{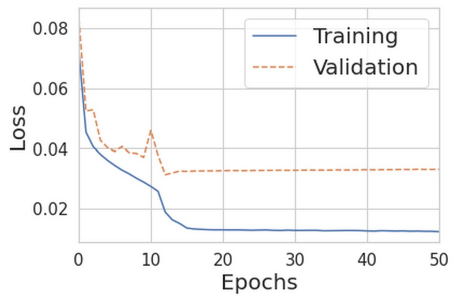} &
\includegraphics[width=0.4\columnwidth]{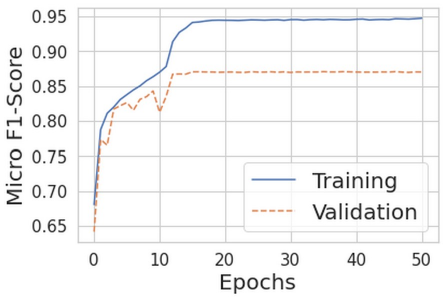}\\
{\scriptsize(a)Loss of EfficientECG on HMIC} & {\scriptsize (b) F1-score of EfficientECG on HMIC}
\end{tabular}
\caption{The training loss and micro-F1 change of EfficientECG on HMIC dataset.}
\label{fig:loss_hmic}
\end{center}
\end{figure}

\subsubsection{Results}
Based on the HMIC dataset, we first conducted a contrastive evaluation on 8-lead ECG data. Figure~\ref{fig:loss_hmic} shows the change in loss during EfficientECG model training. Around the 10th epoch of training, the effect begins to fluctuate. This is due to the changing learning rate. When it reaches the 20th epoch, the model tends to be flat. In contrast, as the learning rate decays, the model still converges until the 50th epoch before triggering the stopping threshold and stopping training. In addition, during the model training process, the changes to micro F1 scores on the validation set are also depicted in Figure~\ref{fig:loss_hmic}.

\begin{figure}[htb]
\centering
\includegraphics[width=\columnwidth]{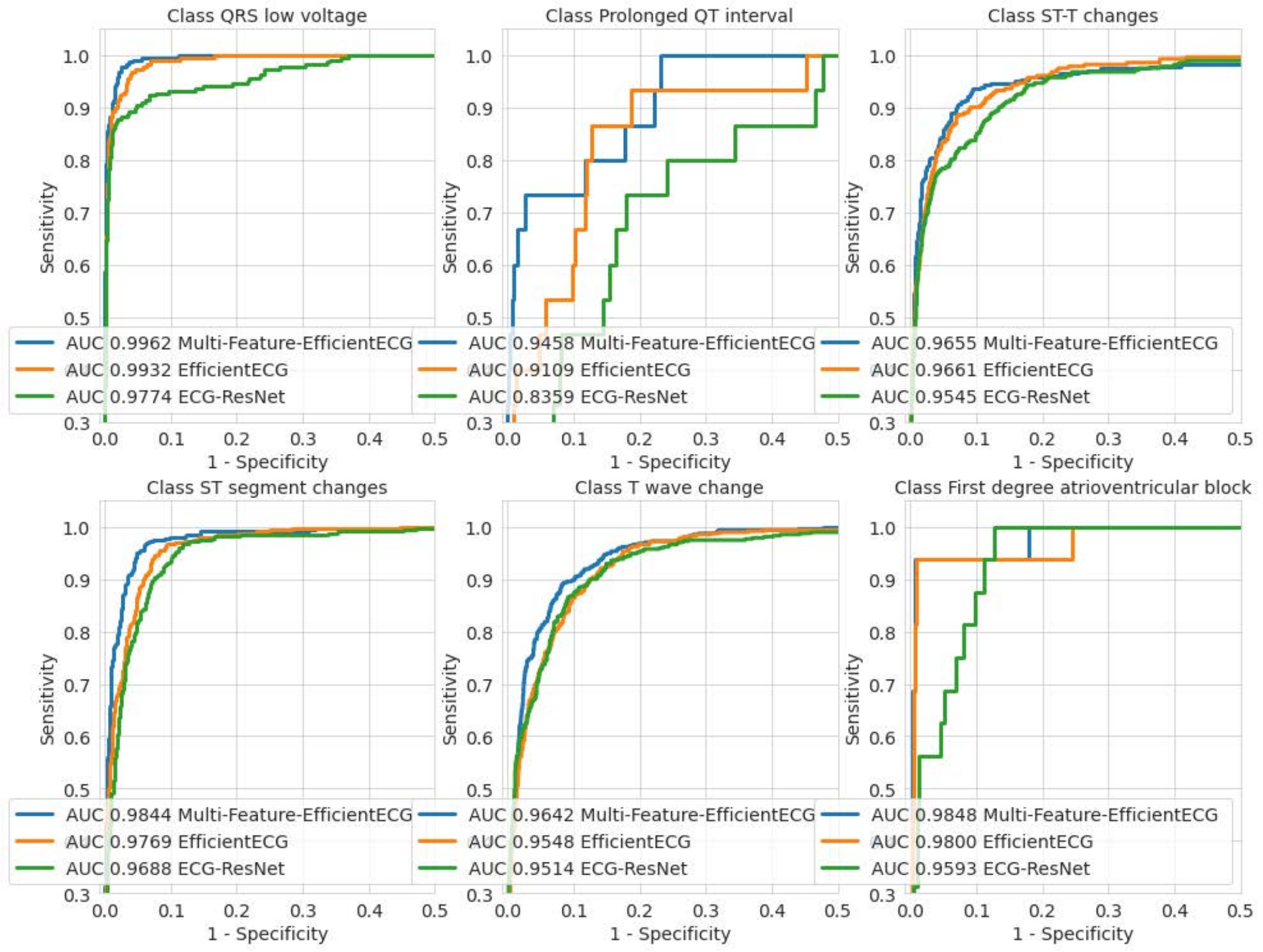}
\caption{The ROC curve and AUC values of different models on HMIC data.}
\label{fig:roc_hmic}
\end{figure}

The evaluation results are shown in Table~\ref{tab:hmic}. The first three rows of data are evaluations conducted on 8-lead ECG data with ECG-ResNet~\cite{Hannun2019}, Transformer~\cite{VaswaniNIPS2017}, and EfficientECG models proposed in this section. It can be seen that EfficientECG uses the fewest parameters, only 42\% of the ResNet, 1.6\% of the Transformer, and 30.1\% of the other EfficientNet, which means EfficientECG consumes fewer computational resources. The parameter numbers difference between EfficientECG and the other EfficientNet is mainly because of the different classification output mapping methods. EfficientECG uses a fully connected layer with sigmoid activation. Nonaka’s EfficientNet uses categories number of prediction blocks, which is more common in multi-output rather than multi-label classification. The micro-F1 values on the training set and the test set are higher than ResNet and Transformer. It further proves that EfficientECG is better than ECG-ResNet in the processing of multi-lead ECG data. The improvement in its classification ability is mainly due to the cross-channel feature fusion ability of the SEBlock module mentioned above. For the embedding layer in the Transformer model, which was originally designed for NLP tasks, it is hard for the Transformer to learn the feature of the ECG data with translation invariance. Although MINA performs slightly better than our EfficientECG on the F1 score during training, it still obtained unsatisfactory results on F1 on the test set. It could be because MINA adopts a relatively larger batch size, and it would converge faster but easily get a local minimum, which contributes to overfitting.

The second group of results are the ablation experiment based on our EfficientECG model, which uses age data, gender data, and cross-attention module as the removal of components respectively. It can be seen from the table that, after adding the gender and age features into the cross-attention module, our EfficientECG obtained the best result on the F1 score. Even though the ablation object without gender data has the highest F1 score on the training set, it still has a lower F1 score on the test set, which would be more likely to lead to overfitting on the simpler data.

To visually compare the classification effect between different models, except the Transformer, for the three models that perform well in the HMIC dataset, we depict the ROC curve with the calculated AUC values in each class from the 8-lead HMIC dataset. Parts of the classes with enough samples are shown in Figure~\ref{fig:roc_hmic}. In most classes, the EfficientECG with additional multi-feature fusion (blue line) has an ROC curve closer to the ideal point on the upper left and a greater AUC value than the EfficientECG without multi-feature and ECG-ResNet models. The graphs demonstrate that the multi-feature fusion EfficientECG model improves the classification effect.

\section{Conclusion and Future Work}
\label{sect:conclusion}

On top of the existing deep learning methods for analysing ECG data, we proposed a novel classification model, EfficientECG, according to the characteristics of ECG data. Subsequently, by proposing a cross-attention module, we enhanced the EfficientECG to a multi-feature fusion model for classifying ECG data including gender, age, and multi-lead ECG data. In comparison with many mainstream models in the field of ECG, classification evaluations were conducted on three representative datasets. Additionally, an ablation study is used to further verify the effect of every component in our multi-feature fusion model. The evaluation results show that our EfficientECG could take advantage of multi-feature data, obtain satisfactory accuracy and consume fewer computational resources. 

Admittedly, there are still some limitations or potential improvements to our work. According to some recent research, even EfficientNet architecture has advantages in parameter count and FLOPs (Floating Point Operations per Second), which indeed reduces the model complexity. However, the inference time is still limited by the memory access of data, which greatly influences the practical application of AI models in real-time ECG diagnosis. Also, the more abundant feature types are worthy to be taken into consideration. In future work, we plan to further investigate how to modify our EfficientECG to fit the more complicated type of additional features in ECG diagnosis, and update the optimization methods in model training and inference to improve the model efficiency and effectiveness.

\bibliographystyle{plain}
\bibliography{reference.bib}

\end{document}